%% file: main.tex
\documentclass[10pt, a4paper, twocolumn]{article} 

\input{structure.tex} 
\usepackage{makecell}
\usepackage{tabularx}
\usepackage{adjustbox}
\usepackage{siunitx}
\usepackage{textgreek}

\title{A one-armed CNN for exoplanet detection from lightcurves} 

\author{
	\authorstyle{Koko Visser\textsuperscript{1}\textsuperscript{*}, Bas Bosma\textsuperscript{2} and Eric Postma\textsuperscript{1}} 
	\newline\newline 
	\textsuperscript{1}\institution{Jheronimus Academy of Data Science, the Netherlands} \\ 
	\textsuperscript{2}\institution{VU Amsterdam, the Netherlands} \\ 
	\textsuperscript{*}\institution{Corresponding author: k.visser@jads.nl}
}

\begin{document}
    \maketitle
    \thispagestyle{firstpage}

    \lettrineabstract{We propose Genesis, a one-armed simplified Convolutional Neural Network (CNN) for exoplanet detection, and compare it to the more complex, two-armed CNN called Astronet. Furthermore, we examine how Monte Carlo cross-validation affects the estimation of the exoplanet detection performance. Finally, we increase the input resolution twofold to assess its effect on performance. The experiments reveal that (i) the reduced complexity of Genesis, i.e., a more than 95\% reduction in the number of free parameters, incurs a small performance cost of about 0.5\% compared to Astronet, (ii) Monte Carlo cross-validation provides a more realistic performance estimate that is almost 0.7\% below the original estimate, and (iii) the twofold increase in input resolution decreases the average performance by about 0.5\%. We conclude by arguing that further exploration of shallower CNN architectures may be beneficial in order to improve the generalizability of CNN-based exoplanet detection across surveys.}

    \section{Introduction}
        CNNs have proven very successful in detecting exoplanets from transit data. They allow for the classification of photometric transit data on a scale and level of accuracy exceeding those of more traditional methods~\citep{MallenOrnelasEtAl03}.
        
        The first successful application of a CNN on transit data was \emph{Astronet} \citep{Shallue18}, with 14 convolutional layers and 4 dense layers. It achieved 96.0\% accuracy in classifying photometric transit data from the Kepler survey. Astronet became a baseline architecture and starting point for further modeling efforts. For instance, \citet{AnsdellEtAl18} proposed \emph{Exonet}, an extension of Astronet that included additional domain knowledge and centroid adjustments. The architecture of Exonet is more complex than Astronet: two extra layers in the input channels and extended dense layers to include stellar parameters. 
        
        The architectures and settings of both Astronet and Exonet came about by considerable optimization and hyperparameter tuning~\citep{Shallue18, AnsdellEtAl18}. Most prominently, both optimized networks incorporate two independent input branches that perform separate convolution operations to process folded lightcurve data on two `views': a \emph{global view} encompassing the entire folded lightcurve and a \emph{local view} dedicated to the processing of the segment where the transit occurs. \citet{Shallue18} argue that the use of two views improves Astronet's accuracy, mainly because the local view emphasizes the properties of the transit itself, whereas the global view entails substantial elements such as secondary eclipses.
                
        However, from a machine learning perspective, the use of two views raises three main concerns. First, part of the global view is replicated in the local view, which introduces redundancy in the processed data. Second, the use of two input channels incurs computational costs. Third, the introduction of two views increases the risk of overfitting \citep{AnsdellEtAl18}, thereby possibly reducing generalizability to transit data from surveys such as K2 and TESS that differ in several respects from those of the Kepler survey~\citep{OsbornEtAl20,  DattiloEtAl19, ChaushevEtAl19}. 
                
        In light of these concerns, we introduce \emph{Genesis}, a reduced-depth single-view CNN, as an alternative baseline architecture (cf. \citet{AnsdellEtAl18}). Apart from having a single view, Genesis contains only four convolutional layers and two dense layers whilst still closely approximating the performance of the more complex dual-view variants.
                
        In evaluating Genesis we performed three experiments. First of all, we compared its performances to those of Astronet. In doing so, we kept as close as we could to the experimental setting of \citet{Shallue18} in terms of procedures and datasets used. The second experiment determines the effect of randomization of data by means of Monte Carlo cross-validation \citep{Picard1984} on the performance estimate. Whilst not reported by \citet{AnsdellEtAl18} and \citet{Shallue18}, we believe that determining the effect of randomization provides important insights into the robustness of results in relation to the complexity of the model. The third experiment determines how enlarging the phase-folded lightcurves affects performance.
                
        The rest of this paper is structured as follows. Section \ref{sec:Exoplanet detection} reviews exoplanet detection with CNNs and characterizes Genesis in the context of mitigating overfitting. We present our methodology to evaluate Genesis in section \ref{sec:methods}, discuss the outcomes of our experiments in section \ref{sec:Results}, and close with a general discussion and conclusion on the virtues of using simpler models in Section \ref{sec:discussion}.

        \section{CNNs and exoplanet detection}
            \label{sec:Exoplanet detection}
            CNNs such as Astronet and Exonet uncover exoplanets based on transit data. Fluctuations in star brightness seen by an observer may exhibit exoplanet appearances (so-called \emph{Threshold-crossing events}, or \emph{TCEs}). These decreases in brightness are often so subtle that pre-processing of the data is required to detect them. In the case of the Kepler survey, lightcurves consist of approximate 30-minute cadences recorded over the course of 17 quarters, yielding about 70,000 observations (brightness samples) per star. Phase folding is used to enhance the signal-to-noise ratio by eliminating the time domain. Phase curves are assembled based on a given periodicity, duration, and start (epoch) in the lightcurves. Finding these parameters of (yet) unknown exoplanets is usually performed by brute force algorithms \citep[cf.][]{KovacsEtAl02} that mark periodic responses of in-transit data. A phase-folded curve can be assembled based on these parameters, typically exhibiting a mix of exoplanet properties, noise, and other phenomena not related to exoplanets. Whether the phase-folded data ultimately characterizes an exoplanet candidate can be predicted with CNNs, and subsequently be (dis)confirmed by follow-up observations. 
    
            To make accurate predictions with CNNs, many factors need to be taken into account. Not only the input data used, but also the architecture of the CNN and the number of trainable parameters in relation to the availability of training data. 
            
\begin{figure*}[!htb]
                \centering
                \includegraphics[scale=0.60]{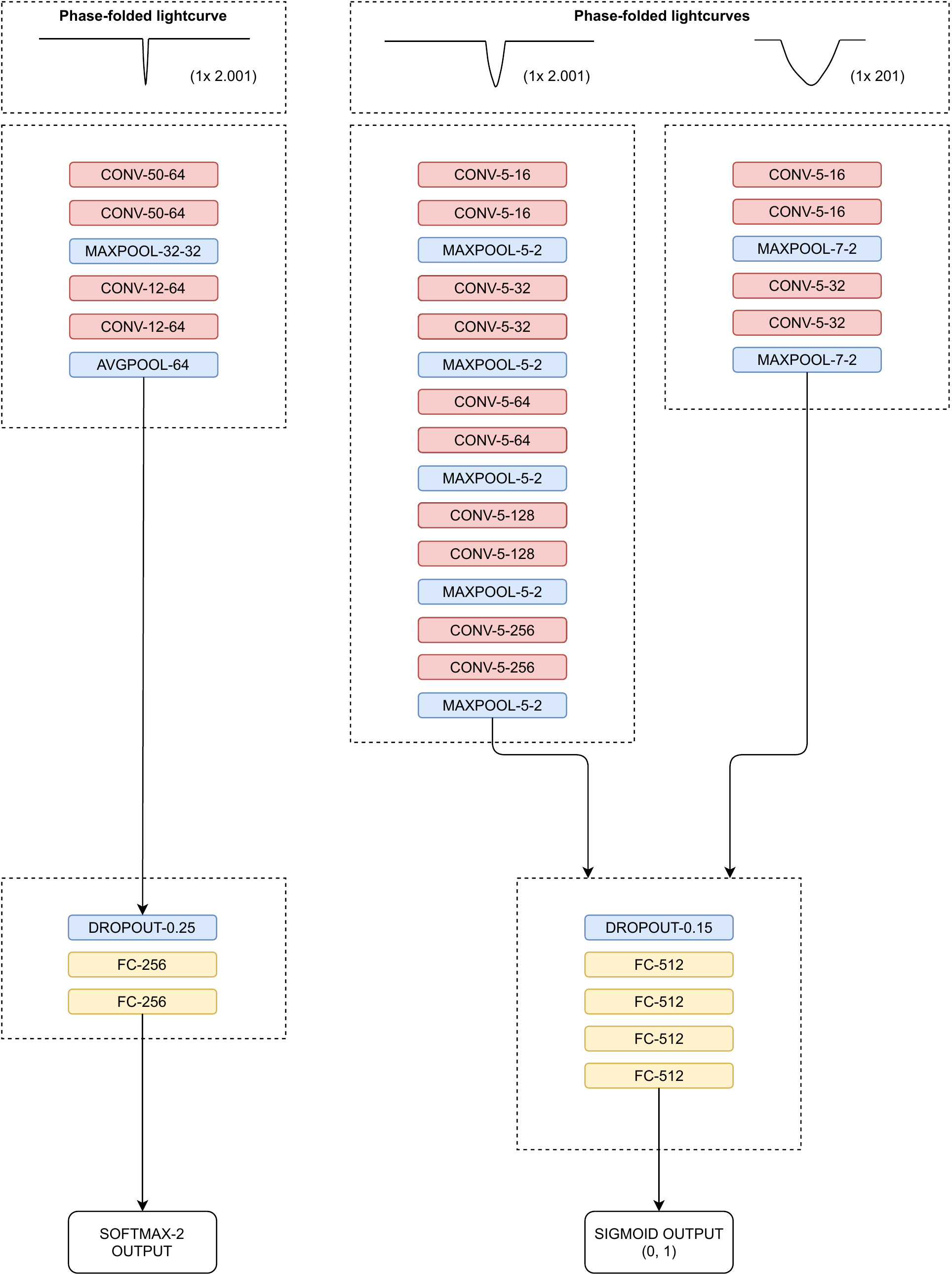}
                \caption{\footnotesize A schematic representation of Genesis (left) and Astronet (right). The three key characteristics of Genesis are: (1) single-view input with 2,001 (or 4,002) input samples instead of dual-view input with 2,001 and 201 samples, (2) reduced number of convolution and dense layers, and (3) two outputs instead of single output. CONV-50-64 represents a convolutional layer of size 50  and 64 is the number of filters. MAXPOOL-32-32 indicates the pooling layer with length 32 and stride 32. AVGPOOL-64 stands for average pooling over 49,216 inputs. DROPOUT-0.25 is a dropout layer with a dropout probability of 25\%. FC-256 is a fully connected layer with 256 hidden neurons.} 
                    \label{fig:Architecture}
                \end{figure*}

            Despite careful specification, even the most successful models show signs of overfitting \citep{XuEtAl18}. To decrease the likelihood of overfitting, Genesis is designed to be much sparser in terms of adaptable parameters than Astronet. Figure~\ref{fig:Architecture} illustrates Genesis (left) alongside Astronet (right). The inputs for both CNNs (phase-folded lightcurves) are shown on top. The sequences of convolution and pooling layers are shown below the inputs. Even from a visual comparison, the reduced complexity of Genesis relative to Astronet is evident. 
       
        \section{Methods}
            \label{sec:methods}
            
            \subsection{Dataset}
                Our experiments rely on two datasets: (i) the original \textit{Astronet dataset} created and provided by \citet{Shallue18}, that was used to train Astronet, and (ii) the \textit{generalized Astronet dataset} created by us. In what follows, we shortly discuss both datasets.
                    
            \subsubsection{Astronet dataset}
                \label{sec:Astronet dataset}
                We downloaded the Astronet dataset from the Github repository\footnote{\url{https://github.com/google-research/exoplanet-ml/tree/master/exoplanet-ml/astronet}} maintained by Shallue. It consists of 15,737 phase-folded lightcurves holding signals of true exo\-planets and non-exoplanets. Each of these phase-folded lightcurves is represented by two 1D vectors: the complete phase-folded lightcurve downsampled to 2,001 bins, and the 201 bins copied from the center of the phase-folded lightcurve. The original lightcurve data \citet{Shallue18} use have been taken from NASA DR24~\citep{Coughlin15}. Each phase-folded lightcurve is labelled with Autovetter~\citep{Cantanzarite15} as either `planet candidate' or `no planet candidate' (\citet{Shallue18} call this groundtruth). The dataset is divided into three parts consisting of approximately 80\% training data, 10\% validation data and 10\% test data. 
                     
            \subsubsection{Generalized Astronet dataset}
                \label{sec:Additional dataset}
                The purpose of the generalized Astronet dataset is twofold: (i) to create phase-folded lightcurves that consist of twice the number of bins compared to those in the Astronet dataset, and (ii) to partition the Astronet dataset into training, validation, and test sets according to the Monte Carlo cross-validation procedure. 
                        
                To construct the generalized Astronet dataset we use the same procedure as used by \citet{Shallue18} for the creation of their Astronet dataset. In addition, we implement the Monte Carlo cross-validation procedure by partitioning the dataset into 20 random training, validation, and test partitions, and prepare the data to allow for a larger binsize.
                
                Using \citet{Lightkurve} we follow a slightly different approach from \citet{Shallue18} in the removal of outliers and the flattening of lightcurve data (i.e., reduction of long-term trends). Using the \texttt{LightCurve.remove\_outliers()} function, we make use of sigma clipping where upper and lower bound parameters are set respectively to {4\textsigma} and {20\textsigma}. Flattening is performed using the Savitzky-Golay filter, that is part of the \texttt{LightCurve.flatten()} function. Finally we use the \texttt{LightCurve.fold()} function to fold the lightcurve data in the desired binsizes.
                
                We create two different global-view binsizes (2,001 and 4,002) resulting in  14,660 and 14,009 lightcurves per dataset, respectively.\footnote{The smaller number of lightcurves for the 4,002 binsize is due to failures to compute the \texttt{LightCurve.flatten()} function.} 
                
            \subsubsection{Data augmentation}
                Data augmentation helps to reduce the risk of overfitting and is implemented, by both horizontal reflections~\citep{Shallue18} and Gaussian noise~\citep{AnsdellEtAl18}, on the Astronet as well as the generalized Astronet dataset. Instead of a single Gaussian version of each lightcurve, as proposed by~\citet{AnsdellEtAl18}, we create four Gaussian lightcurve copies. The random noise is sampled from a normal distribution with mean and standard deviation equal to those of the phase-folded lightcurves in the respective training sets.
                
            \subsection{Architecture and implementation}
                Table~\ref{fig:Comparison-CNNs} specifies the parameters for Genesis, Astronet, and Exonet. Genesis is implemented in Keras \citep{CholletEtAl15} on top of Tensorflow \citep{AbadiEtAl16}. For time efficiency we used hardware acceleration from NVidia (NVidia Titan RTX and NVidia Quaddro) and the cuDNN library \citep{ChetlurEtAl14}. The compute time in our setup to iterate through the training data for instances of Genesis is approximately 285\si{\us} and 465\si{\us} per training step for binsizes 2,001 and 4,002, respectively.
                
                \begin{table}[!htb]
                    \center
                    \begin{adjustbox}{max width=0.49\textwidth}
                        \begin{tabular}{l c c c}
                            
                            \hline
                                & \textbf{Genesis} & \textbf{Astronet} & \textbf{Exonet} \\ [0.5ex]
                            \hline
                                Convolution filter size & 2$\times$50 + 2$\times$12 & 14$\times$5 & 14$\times$5 \\ [0.75ex]
                            \hline
                                Maxpool size & 32 & 2$\times$7 + 5$\times$5 &  2$\times$7 + 5$\times$5  \\ [0.75ex]
                            \hline
                                Stride & 32 & 2 & 2 \\ [0.75ex]
                            \hline
                                Last max/global pool layer & 64 & 7 + 5 & 7 + 5 \\ [0.75ex]
                            \hline
                                Dropout & Yes & Yes & Unknown \\ [0.75ex]
                                \hline
                                    Output transfer function & Softmax & Sigmoid & Sigmoid \\ [0.75ex]
                                \hline
                                    Loss function & \makecell{Categorical \\ cross-entropy} & \makecell{Cross- \\entropy} & \makecell{Cross- \\ entropy} \\ [0.75ex]
                                \hline
                                
                            \end{tabular}
                        \end{adjustbox}
                        
                        \caption{\footnotesize Overview network comparison}
                        \label{fig:Comparison-CNNs}
                    \end{table}    
                
            \subsection{Experimental procedure}
                \citet{Shallue18} evaluated Astronet by means of ensembles of 10 instances.
                In total we create 60 ensembles of 10 instances of Genesis. We train 20 ensembles of Genesis on the Astronet dataset and 20 on both the 2,001-bin and 4,002-bin versions of the generalized Astronet dataset. Each ensemble consists of 10 training sessions. We use Xavier uniform \citep{Glorot10} to initialize the convolutional filters and dense layers, and implement early stopping (with patience = 50 and min delta = 0.1\%). The maximum number of epochs during training is 125. We use the categorical cross-entropy loss function. ReLU is used as activation function. 
                
                For each ensemble, we compute the \emph{ensembled accuracy}. Ensembled accuracy is similar to Accuracy, but instead of calculating it on individual predictions, the arithmetic mean over each TCE prediction made by the 10 trained models on the test data within the ensemble is taken. In addition, we compute the ensembled AUC.
                
        \section{Results}
            \label{sec:Results}

                The first row of table~\ref{Table: Comparison Astronet} shows the performances of Genesis on the Astronet dataset, alongside the input size and number of parameters. For comparison, the second to fourth rows list the results reported for Astronet on the same data \citep{Shallue18}. We make three observations. The first observation is that Genesis performs with an accuracy that is 0.5\% below that of Astronet (second row). The second observation is that Genesis performs 0.1\% better than the Astronet version with a single branch of 2,001 bins (third row). The final observation is that in terms of AUC, Genesis performs worse than all Astronet versions. This reflects the fact that we optimized and selected our models based on accuracy, rather than on AUC, whereas Astronet used AUC as the optimization criterion.
                    
                \begin{table}[!htb]
                    \center
                    \begin{adjustbox}{max width=0.49\textwidth}
                    
                        \begin{tabular}[t]{l r r r r}
                            \multicolumn{5}{c}{}\\
                            \hline
                                
                            \textbf{architecture} & \textbf{input size} & \textbf{parameters} & \textbf{ACC} & \textbf{AUC} \\ [0.5ex] 
                                               \hline
                                               
                            Genesis & 2,001 & 0.389M  & 95.5\% & 94.5\% \\ [0.75ex]
                            \hline \hline
        
                            Astronet & 2,001 and 201 & >8.793M & 96.0\% & 98.8\% \\ [0.75ex]
                            \hline                
                            
                            Astronet\textsubscript{global} & 2,001 & >8.128M & 95.4\% & 98.5\% \\ [0.75ex]
                            \hline                
              
                            Astronet\textsubscript{local} & 201 & >1.453M & 92.4\% & 97.3\% \\ [0.75ex]
                            \hline                
                        \end{tabular}
                    
                    \end{adjustbox}
                        
                    \caption{\footnotesize Comparative evaluation of the performance obtained with Genesis and Astronet. The parameters for the individual models and branches were based on the `trainable parameters' reported by Keras after we constructed the CNNs. Ensembled accuracy (ACC) and ensembled AUC (AUC) for Astronet are provided in \citet{Shallue18}. Reperformance by \citet{AnsdellEtAl18} indicates a similar ACC for Astronet: 95.8\%}
                    \label{Table: Comparison Astronet}
                \end{table}
                 
            \subsection{Monte Carlo cross-validation}
\begin{figure*}
                    \centering
                    \begin{minipage}[b]{.45\textwidth}
                        \includegraphics[scale=0.4]{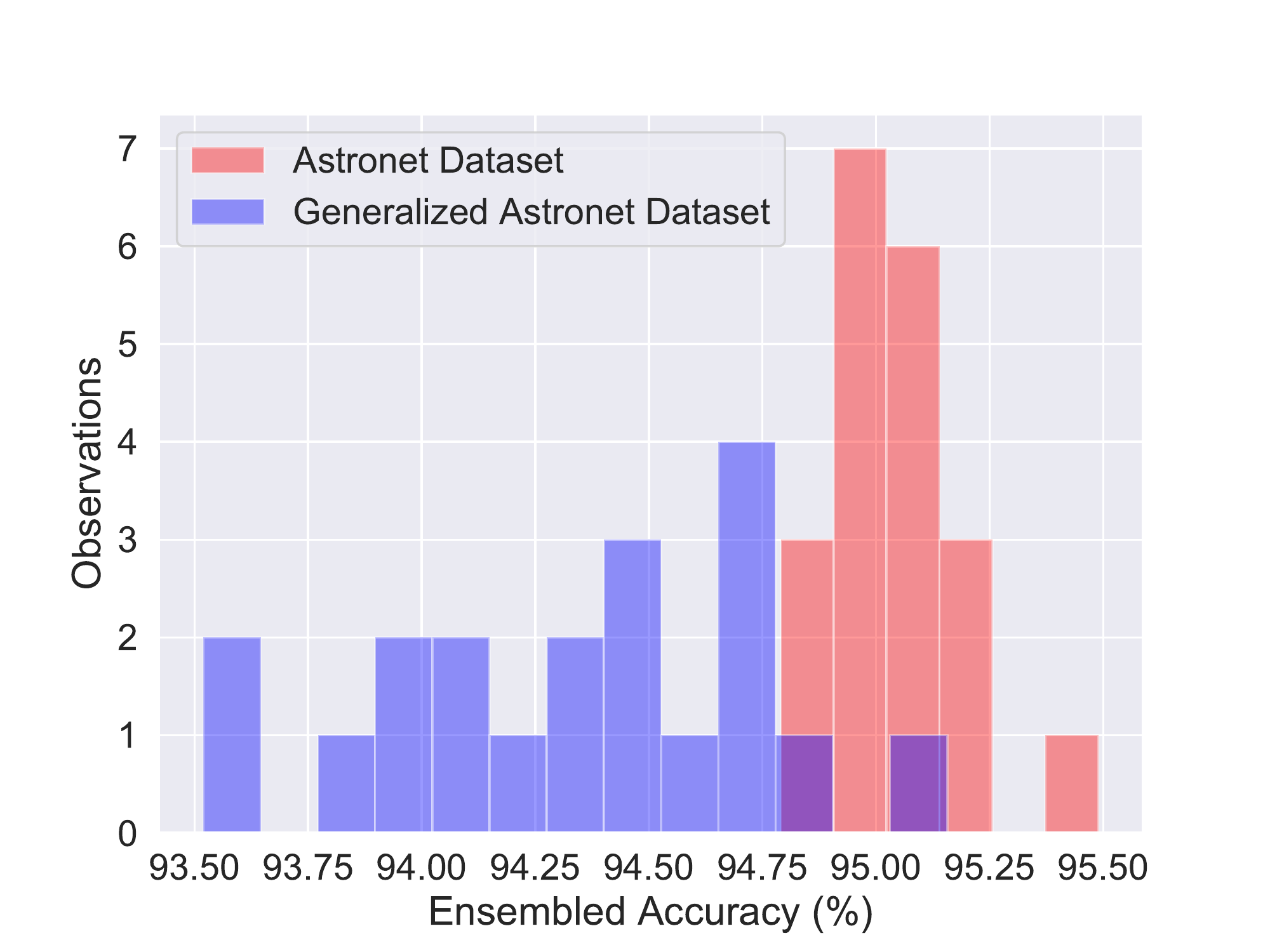}
                        \caption{\footnotesize Histogram of accuracies obtained with Genesis over 100 replications on the Astronet dataset (red) and generalized Astronet dataset holding 2,001 bins (blue).} 
                        \label{fig:datarandom}
                    \end{minipage}\qquad
                    \begin{minipage}[b]{.45\textwidth}
                        \includegraphics[scale=0.4]{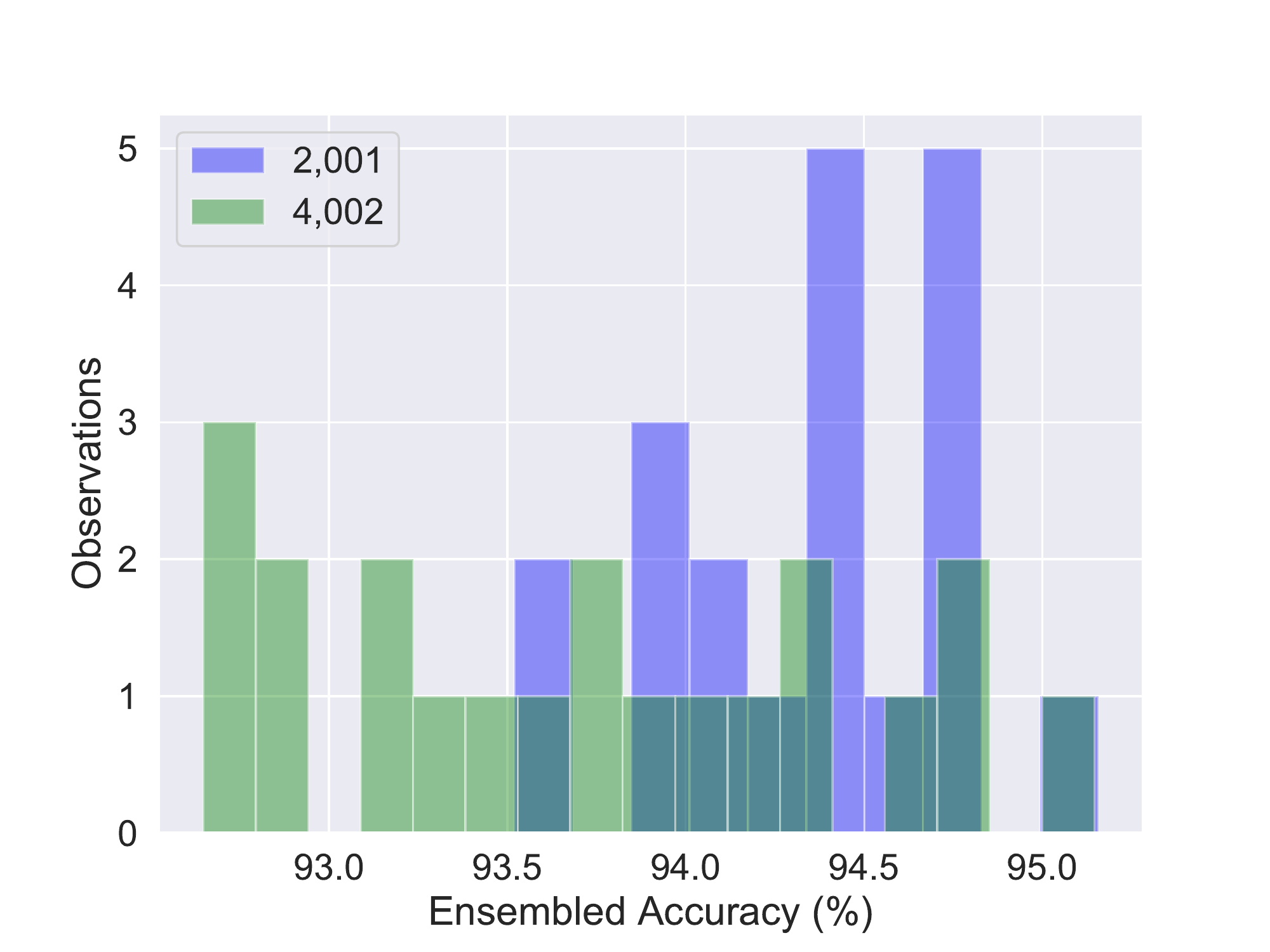}
                        \caption{\footnotesize Histogram of accuracies obtained with Genesis over 100 replications on the two variants of the generalized Astronet dataset containing 2,001 (blue) and 4,002 (green) bins.} 
                        \label{fig:datarandom_2001_4002}
                    \end{minipage}
                \end{figure*}

                The results of training Genesis on the 2,001-bin version of the generalized Astronet dataset are shown in Figure~\ref{fig:datarandom}. The histograms represent the distributions of ensembled accuracies obtained with (blue) and without (red) randomization. Whereas the distributions are clearly different, their means differ by only 0.5\%. The specific choice of training, validation, and test set for the Astronet dataset, results in a small (0.7\%) overestimate of the performance.
                
            \subsection{Larger input size}
                The results of training Genesis on both variants of the generalized Astronet datasets are listed in Figure~\ref{fig:datarandom_2001_4002}. The performances of the 2,001-bin and 4,002-bin versions reveal that using more bins does not result in higher ensembled accuracies (maximum ensembled accuracy 95.5\% vs. 95.1\% and average ensembled accuracy of 94.3\% vs. 93.8\%). Besides, increasing the number of bins leads to a higher variance.
                
        \section{Discussion and conclusions}
            \label{sec:discussion}
            The general tenet of deep learning is to create deeper networks to improve performance \citep{Szegedy2014,He2015}. At the same time, however, we know that the performance of deep networks can be closely approximated using shallower ones \citep[e.g.,][]{Ba2013}. Our findings reveal that a more than 95\% reduction in the number of parameters incurs only a small cost in prediction performance. In addition, we found that despite the large size of the Astronet dataset, Monte Carlo cross-validation provides a slightly different and more reliable estimate of the predictive performance. Increasing the input size by two incurs a considerable computational cost and does not offer any performance improvement. \\
        
            \noindent Using more shallow CNN architectures offers the potential benefit of improved generalization. Our results and their considerations lead us to conclude that further exploration of simpler CNN architectures may be beneficial to the generalizability of exoplanet detection across surveys.   

    \printbibliography 

 
\end{document}

%% file: structure.tex
\usepackage[english]{babel} 
\usepackage{microtype} 
\usepackage{amsmath,amsfonts,amsthm} 
\usepackage[svgnames]{xcolor} 
\usepackage[hang, small, labelfont=bf, up, textfont=it]{caption} 
\usepackage{booktabs} 
\usepackage{lastpage} 
\usepackage{graphicx} 
\usepackage{enumitem} 
\setlist{noitemsep} 
\usepackage{sectsty} 

\allsectionsfont{\usefont{OT1}{phv}{b}{n}} 

\usepackage{geometry} 

\usepackage[T1]{fontenc} 
\usepackage[utf8]{inputenc} 

\usepackage{XCharter} 

\usepackage{fancyhdr} 
\fancypagestyle{firstpage}{ 
	\fancyhf{}
}

\newcommand{\authorstyle}[1]{{\large\usefont{OT1}{phv}{b}{n}\color{DarkRed}#1}} 

\newcommand{\institution}[1]{{\footnotesize\usefont{OT1}{phv}{m}{sl}\color{Black}#1}} 

\usepackage{titling} 

\newcommand{\HorRule}{\color{DarkGoldenrod}\rule{\linewidth}{1pt}} 

\pretitle{
	\vspace{-30pt} 
	\HorRule\vspace{10pt} 
	\fontsize{32}{36}\usefont{OT1}{phv}{b}{n}\selectfont 
	\color{DarkRed} 
}

\posttitle{\par\vskip 15pt} 

\preauthor{} 

\postauthor{ 
	\vspace{10pt} 
	\par\HorRule 
	\vspace{20pt} 
}

\usepackage{lettrine} 
\usepackage{fix-cm}	

\newcommand{\initial}[1]{ 
	\lettrine[lines=3,findent=4pt,nindent=0pt]{
		\color{DarkGoldenrod}
		{#1}
	}{}%
}

\usepackage{xstring} 

\newcommand{\lettrineabstract}[1]{
	\StrLeft{#1}{1}[\firstletter] 
	\initial{\firstletter}\textbf{\StrGobbleLeft{#1}{1}} 
}

\usepackage[backend=bibtex,style=authoryear,natbib=true]{biblatex} 

\usepackage[autostyle=true]{csquotes} 